%% file: avatar_camera_ready.tex
\documentclass[letterpaper]{article} % DO NOT CHANGE THIS
\usepackage{aaai2026}  % DO NOT CHANGE THIS
\usepackage{times}  % DO NOT CHANGE THIS
\usepackage{helvet}  % DO NOT CHANGE THIS
\usepackage{courier}  % DO NOT CHANGE THIS
\usepackage[hyphens]{url}  % DO NOT CHANGE THIS
\usepackage{graphicx} % DO NOT CHANGE THIS
\usepackage{natbib}  % DO NOT CHANGE THIS AND DO NOT ADD ANY OPTIONS TO IT
\usepackage{caption} % DO NOT CHANGE THIS AND DO NOT ADD ANY OPTIONS TO IT
\usepackage{algorithm}
\usepackage{algorithmic}

\usepackage{newfloat}
\usepackage{listings}
\DeclareCaptionStyle{ruled}{labelfont=normalfont,labelsep=colon,strut=off} % DO NOT CHANGE THIS
\floatstyle{ruled}
\newfloat{listing}{tb}{lst}{}
\floatname{listing}{Listing}
\title{MonoCloth: Reconstruction and Animation of Cloth-Decoupled Human Avatars from Monocular Videos}
\author{
    Daisheng Jin,
    Ying He\thanks{Corresponding author}
}
\affiliations{
    S-Lab, Nanyang Technological University, Singapore\\
    daisheng001@e.ntu.edu.sg, YHe@ntu.edu.sg
}

\usepackage{bibentry}
\input{AnonymousSubmission/LaTeX/preamble}

\begin{document}

\maketitle

\begin{figure*}[t]
    \centering
    \includegraphics[width=0.95\linewidth]{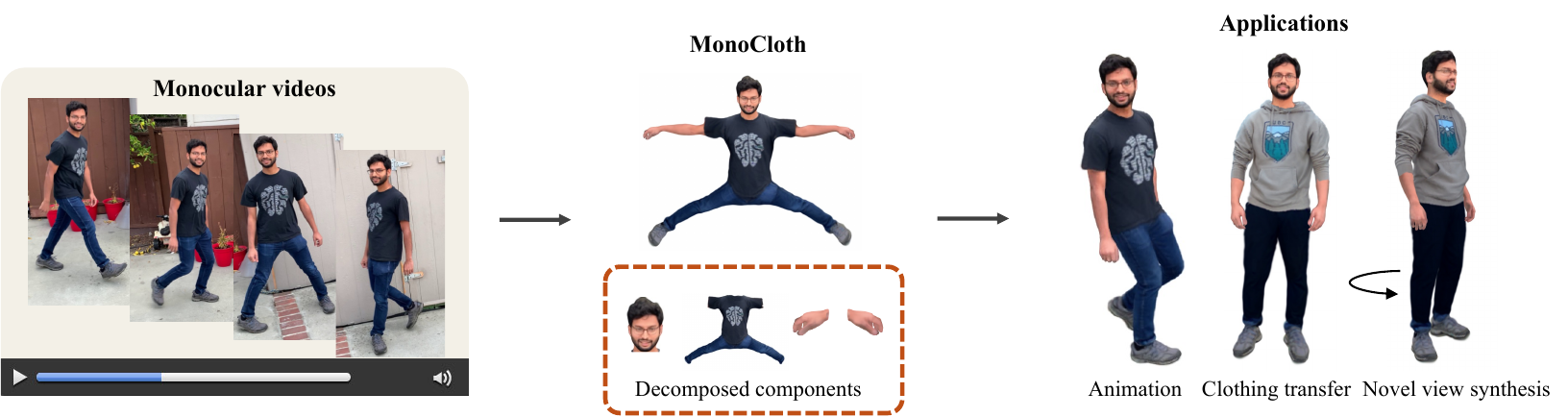}
    \captionof{figure}{\textbf{MonoCloth} reconstructs a human avatar from monocular videos by dividing it into separate components. Each component is optimized using a strategy suited to its geometric and motion characteristics, which improves reconstruction quality. The reconstructed avatar supports natural animation and can be rendered from novel viewpoints. The modular design also allows for part-level editing, such as clothing transfer.}
    \label{fig:teaser}
    % \vspace{-10pt}
\end{figure*}

\input{AnonymousSubmission/LaTeX/secs/0_abstract}

\begin{links}
    \link{Project}{https://kingjg.github.io/MonoCloth.github.io}
\end{links}

\input{AnonymousSubmission/LaTeX/secs/1_intro}
\input{AnonymousSubmission/LaTeX/secs/2_related}

% \input{secs/3_preliminary}
\input{AnonymousSubmission/LaTeX/secs/4_method}
\input{AnonymousSubmission/LaTeX/secs/5_experiments}
\input{AnonymousSubmission/LaTeX/secs/6_conclusion}

\section*{Acknowledgments}
This work was supported in part by the Ministry of Education, Singapore, under its Academic Research Fund Grant (RT19/22), as well as cash and in-kind funding from NTU S-Lab and the industry partner(s).

\bibliography{AnonymousSubmission/LaTeX/avatar_ref}

\end{document}

%% file: AnonymousSubmission/LaTeX/preamble.tex
\usepackage{amsmath}
\usepackage{amsfonts}
\usepackage{booktabs}
\usepackage[table]{xcolor}
\usepackage{dashbox}
\usepackage[framemethod=tikz]{mdframed}
\usepackage{xcolor}
\usepackage{graphicx}
\usepackage{rotating}
\usepackage{multirow}
\usepackage{pifont}

\usepackage{cuted}

\newmdenv[hidealllines=false,roundcorner=5pt]{linebox}

\graphicspath{{low_hes/}}

%% file: AnonymousSubmission/LaTeX/secs/0_abstract.tex
\begin{abstract}
% We present \textbf{MonoCloth}, a novel method for reconstructing and animating clothed human avatars from monocular videos. Although monocular videos provide rich visual information, they lack full 3D geometry, which makes it difficult to recover body shape and motion accurately. This gap limits both the quality of reconstruction and the realism of animations. To address this, we adopt a part-based decomposition strategy that separates the human avatar into body, face, hands, and clothing. This design reflects the varying levels of reconstruction difficulty and deformation complexity in different regions. We focus on a detailed geometry recovery for the face and hands. For clothing, we develop a cloth motion simulation module that captures garment deformation by incorporating temporal motion cues and geometric information. Our experiments show that \textbf{MonoCloth} improves both the visual quality of reconstruction and the realism of animation compared to existing methods. Thanks to its part-based design, \textbf{MonoCloth} also supports additional tasks such as clothing transfer, making it useful for a wide range of applications.
Reconstructing realistic 3D human avatars from monocular videos is a challenging task due to the limited geometric information and complex non-rigid motion involved. We present \textbf{MonoCloth}, a new method for reconstructing and animating clothed human avatars from monocular videos. To overcome the limitations of monocular input, we introduce a part-based decomposition strategy that separates the avatar into body, face, hands, and clothing. This design reflects the varying levels of reconstruction difficulty and deformation complexity across these components. Specifically, we focus on detailed geometry recovery for the face and hands. For clothing, we propose a dedicated cloth simulation module that captures garment deformation using temporal motion cues and geometric constraints. Experimental results demonstrate that MonoCloth improves both visual reconstruction quality and animation realism compared to existing methods. Furthermore, thanks to its part-based design, MonoCloth also supports additional tasks such as clothing transfer, underscoring its versatility and practical utility.
\end{abstract}

%% file: AnonymousSubmission/LaTeX/secs/1_intro.tex
\section{Introduction}

3D human avatars are used in many areas, including film production, game development, and AR/VR applications. However, creating high-quality avatars often requires manual modeling or expensive 3D capture setups, which limits their use in everyday settings.

To make avatar creation more accessible, recent work has focused on using low-cost inputs such as images and videos. 
The development of implicit neural representations has made it possible to reconstruct both 3D geometry and appearance from 2D images~\cite{mildenhall2021nerf,jin2025sfdm}. 
When applied to multi-view videos, these methods enable the reconstruction of high-quality deformable 3D human avatars. 
For example, Animatable NeRF~\cite{peng2021animatable} and AvatarRex~\cite{zheng2023avatarrex} build implicit fields in a canonical space and animate them using deformation functions. Although effective, these methods require synchronized multi-view capture, which is difficult to set up outside of controlled environments.

% To lower the cost of input data, recent methods have extended avatar reconstruction to monocular video. 
To reduce input cost, some recent methods focus on reconstructing avatars from monocular videos.
HumanNeRF~\cite{weng2022humannerf} maps frames with different poses to a unified T-pose space and learns an implicit neural field for rendering from new viewpoints.
InstantAvatar~\cite{jiang2023instantavatar} improves training and rendering speed by using Instant-NGP~\cite{muller2022instant}, an efficient NeRF backbone. 
Instead of implicit fields, 3DGS-Avatar~\cite{qian20243dgs} represents avatars using 3D Gaussian primitives~\cite{kerbl20233d}, which improves both efficiency and visual quality. 
ExAvatar~\cite{moon2024expressive} adds detailed modeling of facial expressions and hand articulations to enrich avatar expressiveness.   Vid2Avatar-Pro~\cite{guo2025vid2avatarpro} uses pretraining on a large multi-view video dataset to help the model learn useful priors for reconstruction from monocular input. 

% However, most existing methods for monocular avatar reconstruction mainly focus on detailed appearance modeling, while paying limited attention to the avatar's behavior during animation. 
% As a result, animations often exhibit noticeable frame-to-frame inconsistencies and unnatural appearance variations due to an insufficient understanding of temporal dynamics in geometry and appearance.
% This dynamic mismatch also hinders the model’s ability to align inter-frame variations with a consistent avatar representation, ultimately limiting the reconstruction quality, especially for fine details.
% To address these limitations, we propose \textbf{MonoCloth}, a novel framework that improves the model’s ability to capture human dynamics from monocular videos, enabling more consistent and accurate avatar reconstruction.

While these methods focus on detailed appearance modeling, they often overlook how the avatar moves over time. As a result, animations may show inconsistencies between frames or unnatural changes in appearance. This happens because the model does not fully capture how shape and appearance evolve across time, making it harder to maintain a stable avatar and recover fine details.

% \textbf{MonoCloth} builds on the 3D human parametric model SMPL-X~\cite{pavlakos2019smplx} and decomposes the human avatar into two components (body and outfit) based on the observation that different regions exhibit varying degrees of rigidity and deformation. 
% The body primarily undergoes articulated, rigid transformations driven by joint rotations, while clothing, typically soft and loosely fitted, experiences complex non-rigid deformations influenced by both body motion and secondary physical effects such as inertia and gravity.
To address these challenges, we present \textbf{MonoCloth}, a method designed to better model human motion from monocular videos, leading to more consistent and accurate reconstructions. 
% MonoCloth uses the parametric body model SMPL-X~\cite{pavlakos2019smplx} as a canvas and separates the avatar into two main parts: the body and the clothing. 
Building upon the parametric body model SMPL-X~\cite{pavlakos2019smplx}, MonoCloth decomposes the human avatar into two major components: the body and the clothing.
The body moves in mostly rigid, joint-driven ways, while clothing undergoes soft, non-rigid motion influenced by physical effects such as inertia and gravity. 

% To effectively model this disparity, we independently optimize the two components. For the bare body, where non-rigid dynamics are limited, we enhance reconstruction quality by leveraging priors from parametric models for the face and hands. For clothing, we introduce a cloth motion simulation module (\textbf{CloSim}) to capture dynamic garment behavior.

We optimize these two parts separately to account for their different behaviors. For the body, where motion is more rigid, we improve reconstruction using strong priors from face and hand parametric models. 
For clothing, we introduce a cloth simulation module (\textbf{CloSim}), which models how garments move during motion. To train CloSim, we first use 2D clothing segmentation to decouple the garment region. Instead of using only a single pose frame as input, we feed in a sequence of poses to capture motion over time. Clothing and motion features are passed through a graph convolutional network (GCN)~\cite{kipf2016semi} to collect local spatial information, followed by a lightweight gated recurrent unit (GRU)~\cite{chung2014empirical} to learn how the motion evolves across frames. This setup helps the model learn both spatial and temporal effects in garment deformation. 

Because modeling clothing dynamics from monocular input is difficult, we use additional information to guide the model. Specifically, we apply vision foundation models to predict depth and surface normals from the video frames. These serve as rough 3D cues during training. We also pretrain on monocular videos of different people to learn general patterns in appearance and clothing motion. With this design, MonoCloth reconstructs high-quality avatars from monocular videos and supports realistic animation under new motion sequences.

Our main contributions are summarized as follows:

\begin{itemize}

% \item We propose a novel part-based Gaussian avatar decomposition, enabling flexible and targeted optimization strategies for different body components.
\item We propose a part-based avatar design built on SMPL-X and 3D Gaussians, allowing targeted optimization for each component.

% \item Based on our explicitly structured avatar, we introduce a cloth motion simulation module that models temporal continuity and spatial consistency of flexible garments, resulting in more physically plausible animations.
% \item Based on this design, we introduce a cloth simulation module that learns both spatial and temporal garment behavior using physically inspired signals.
\item Based on this design, we introduce a cloth simulation module that integrates spatial and temporal cues, aiming to produce more physically consistent clothing motion.

\item Extensive experiments demonstrate that MonoCloth improves both reconstruction quality and animation consistency. The part-based design also supports additional tasks such as clothing transfer.

\end{itemize}

%% file: AnonymousSubmission/LaTeX/secs/2_related.tex
\section{Related Work}

\subsection{3D Human Avatar}

Recent research has explored the reconstruction of 3D human avatars from a variety of input sources, including multi-view videos~\cite{lin2024layga,zhang2025disentangled}, monocular videos~\cite{song2025ctrlavatar}, and single images~\cite{pang2025disco4d,qiu2025lhm}.

Multi-view video-based methods typically rely on controlled laboratory environments with ideal lighting, synchronized camera setups, and known calibration parameters, providing rich and accurate 3D information.
With the advancement of multi-view 3D reconstruction techniques, methods such as AvatarRex~\cite{zheng2023avatarrex} and AnimatableGaussian~\cite{li2024animatable} have adopted Neural Radiance Fields (NeRF)~\cite{mildenhall2021nerf} and 3D Gaussian splatting~\cite{kerbl20233d}, respectively, to reconstruct detailed human avatars. 
% Building on this foundation, these methods further incorporate dynamic motion patterns learned from video sequences, enabling the generation of fully animatable 3D avatars.
Building on these foundations, they further learn dynamic motion patterns from video sequences, enabling the creation of fully animatable 3D avatars.
In contrast, single-image reconstruction offers a more accessible and cost-effective solution but presents significant challenges in recovering fine-grained geometry and maintaining 3D consistency~\cite{zhang2024mimicmotion,hu2024animate}. 

% Multi-view video-based methods typically rely on controlled laboratory environments with ideal lighting, synchronized cameras, and known calibration parameters, providing rich and accurate 3D information.
% With advances in multi-view 3D reconstruction, approaches such as AvatarRex~\cite{zheng2023avatarrex} and AnimatableGaussian~\cite{li2024animatable} leverage Neural Radiance Fields (NeRF)~\cite{mildenhall2021nerf} and 3D Gaussian splatting~\cite{kerbl20233d} to reconstruct highly detailed human avatars.
% Building on these foundations, they further learn dynamic motion patterns from video sequences, enabling the creation of fully animatable 3D avatars.
% In contrast, single-image reconstruction is more accessible and cost-effective but faces substantial challenges in recovering fine-grained geometry and ensuring 3D consistency~\cite{zhang2024mimicmotion,hu2024animate}.

By contrast, in-the-wild monocular videos are convenient to obtain and naturally contain more 3D cues and motion information.
% However, as the input remains fundamentally 2D, models still face significant challenges in accurately interpreting 3D structures.
However, as the input is inherently 2D, accurately inferring 3D structures remains challenging.
GaussianAvatar~\cite{hu2024gaussianavatar} and HUGS~\cite{kocabas2024hugs} address this by representing avatars using 3D Gaussians and introducing learnable offsets to model appearance variations associated with human motion.
ExAvatar~\cite{moon2024expressive} tackles the common limitation of prior works that often neglect facial expressions and hand movements by explicitly modeling these regions, resulting in significantly enhanced avatar realism.
% Vid2AvatarPro~\cite{guo2025vid2avatarpro} and PGHM~\cite{peng2025parametric} compensate for the limited 3D information in monocular videos by pretraining a prior model on multi-view video data, thereby improving both the 3D perception and robustness of the system.
Vid2Avatar-Pro~\cite{guo2025vid2avatarpro} and PGHM~\cite{peng2025parametric} enhance 3D perception and robustness by pretraining on multi-view video data to compensate for the limited 3D cues in monocular inputs.
Nevertheless, most monocular video-based methods treat the outfit and human body as a unified whole during optimization, overlooking their distinct motion patterns. 
% In contrast, our method explicitly accounts for the unique dynamics of garments. We propose a dedicated CloSim module that models clothing motion from both spatial and temporal perspectives, leading to more accurate and detailed avatar reconstructions.
In contrast, our method explicitly models garment dynamics through a dedicated CloSim module that captures clothing motion across both spatial and temporal dimensions, enabling more accurate and detailed avatar reconstructions.

% In contrast, in-the-wild monocular videos are easy to capture and naturally contain more 3D cues and motion dynamics.
% However, as the input is inherently 2D, accurately inferring 3D structures remains challenging.
% GaussianAvatar~\cite{hu2024gaussianavatar} and HUGS~\cite{kocabas2024hugs} address this by representing avatars with 3D Gaussians and introducing learnable offsets to capture motion-related appearance variations.
% ExAvatar~\cite{moon2024expressive} tackles the common limitation of prior works that often neglect facial expressions and hand movements by explicitly modeling these regions, resulting in significantly enhanced avatar realism.
% Vid2AvatarPro~\cite{guo2025vid2avatarpro} and PGHM~\cite{peng2025parametric} enhance 3D perception and robustness by pretraining on multi-view video data to compensate for the limited 3D cues in monocular inputs.
% Yet, most monocular approaches still optimize clothing and body as a single entity, neglecting their distinct motion patterns.
% In contrast, our method explicitly models garment dynamics through a dedicated CloSim module that captures clothing motion across both spatial and temporal dimensions, enabling more accurate and detailed avatar reconstructions.

\begin{table}[t]
\centering
\resizebox{0.45\textwidth}{!}{
\renewcommand{\arraystretch}{1.2}
\begin{tabular}{c|cc|cc|cc}
\hline
\multirow{2}{*}{\textbf{Method}} & \multicolumn{2}{c|}{\textbf{Inputs}} & \multicolumn{2}{c|}{\textbf{Design}} & \multicolumn{2}{c}{\textbf{Outputs}} \\
% \cline{2-6}
& \textbf{Vid.} & \textbf{Pre.} & \textbf{Temp.} & \textbf{Decomp.} & \textbf{Anim.} & \textbf{Trans.} \\
\hline
\hline
% LayerAvatar~\cite{} & Multi-view & None & \ding{51} & \ding{51} & \ding{51} \\
AvatarRex & Multi & None & \ding{55} & \ding{55} & \ding{51} & \ding{55} \\
PhysAvatar & Multi & None & \ding{51} & \ding{51} & \ding{51} & \ding{51} \\
\hline
D$^3$-Human & Mono & None & \ding{51} & \ding{51} & \ding{55} & \ding{51} \\
% GaussianAvatar & Mono & None & \ding{55} & \ding{55} & \ding{51} & \ding{55} \\
ExAvatar & Mono & None & \ding{55} & \ding{55} & \ding{51} & \ding{55} \\
% PGHM & Mono & Multi & \ding{55} & \ding{55} & \ding{51} & \ding{55} \\
Vid2Avatar-Pro & Mono & Multi & \ding{55} & \ding{55} & \ding{51} & \ding{55} \\
\textbf{Ours} & Mono & Mono & \ding{51} & \ding{51} & \ding{51} & \ding{51} \\
\hline
\end{tabular}
}

\caption{\textbf{Comparison of video-based human avatar reconstruction methods.} We compare each method based on input type (Vid.), use of pretraining (Pre.), temporal modeling (Temp.), clothing decomposition (Decomp.), animation support (Anim.), and ability of clothing transfer (Trans.).}
\end{table}

% Conversely, approaches like Disco4D utilize the SMPL-X parametric model combined with 3D Gaussian representations to explicitly reconstruct avatars. Nevertheless, they often depend on synthesized multi-view data, which can result in suboptimal side-view quality.

% Monocular video methods leverage conveniently captured in-the-wild videos, rich in dynamic information but inherently limited to 2D data, complicating accurate 3D interpretation. GaussianAvatar addresses this through 3D Gaussian representations and offset adjustments for appearance variation. ExAvatar improves upon previous techniques by explicitly modeling detailed facial expressions and hand motions, enhancing overall realism. Vid2AvatarPro mitigates inherent monocular limitations by incorporating multi-view video pretrained priors, significantly enhancing robustness and 3D awareness.
% However, most monocular video-based methods treat garments and the human body as a unified entity, neglecting their distinct motion patterns. To address this, our approach introduces a dedicated CloSim module that specifically models garment dynamics from spatial and temporal dimensions, enabling more accurate and finely-detailed avatar reconstruction.

\subsection{Clothing Reconstruction and Motion Simulation}

% Clothing reconstruction plays a critical role in clothed human avatar modeling, as garments typically cover the majority of the body and largely determine the overall rendering quality of the avatar. However, clothing geometry and motion are highly complex due to significant non-rigid deformations during movement, resulting in rich variations in both shape and appearance. As a result, a substantial body of work has focused on reconstructing garment geometry.

% Garment4D reconstructs detailed cloth geometry from 3D point clouds, which inherently encode spatial structure. GaussianGarments leverages multi-view videos and represents garments using 3D Gaussians, enabling the joint reconstruction of geometry and appearance from multi-view visual input. PhysAvatar goes beyond garment modeling and reconstructs full dressed humans. Also relying on multi-view videos, it achieves high-fidelity geometry and appearance, and further incorporates physical parameters of fabric to produce more realistic and smooth cloth animation.

Clothing reconstruction is essential for clothed human avatar modeling, as garments often cover most of the body. However, the complex, non-rigid deformations of clothing during motion lead to significant challenges in modeling its geometry and appearance.
To address this, many studies focus on garment reconstruction. Garment4D~\cite{hong2021garment4d} recovers fine cloth geometry directly from 3D point clouds, leveraging their rich 3D information. GaussianGarments~\cite{rong2024gaussian_garments} uses multi-view videos and 3D Gaussians to jointly reconstruct garment geometry and appearance. PhysAvatar~\cite{zheng2024physavatar} extends beyond garments to reconstruct dressed humans with high-fidelity geometry and appearance, incorporating physical fabric parameters to produce more realistic cloth dynamics.

In contrast, monocular videos provide limited 3D information, making accurate reconstruction significantly more challenging. Typically, recent works aim to infer human geometry directly from 2D input. 
% For example, Reloo~\cite{guo2024reloo} introduces a non-hierarchical virtual bone deformation module to model the motion of loose garments, enabling high-quality prediction of human normal maps from monocular videos. 
Reloo~\cite{guo2024reloo} models loose clothing with virtual bone deformation for high-quality monocular normal prediction.
D$^3$-Human~\cite{chen2025d3human} reconstructs full-body garment geometry from monocular input and supports dynamic cloth simulation during animation. While these methods mainly focus on geometric reconstruction, achieving photorealistic rendering still requires deeper integration of appearance modeling.

In this work, we propose a method for reconstructing high-quality, animatable avatars from low-cost monocular videos. By explicitly decoupling body components and introducing a cloth-specific motion model, our approach optimizes Gaussian avatars to directly render dynamic human animations with fine geometric and appearance details.

% In contrast, monocular videos offer very limited 3D information, making the reconstruction task significantly more challenging. Many recent works focus on predicting human geometry directly from 2D input. Reloo proposes a non-hierarchical virtual bone deformation module to model the motion of loose clothing, enabling high-quality prediction of human normal maps from monocular video. D3 Human reconstructs full-body garment geometry from monocular input and can simulate dynamic clothing motion during animation. While these approaches focus primarily on cloth geometry reconstruction, further integration of appearance modeling is required to achieve photorealistic rendering.

% In this work, we propose a method that reconstructs high-quality animatable avatars from low-cost monocular video. By explicitly decoupling body components and incorporating cloth-specific motion modeling, our method optimizes Gaussian avatars to support direct rendering of dynamic human animation sequences with rich geometry and appearance details.

%% file: AnonymousSubmission/LaTeX/secs/4_method.tex
\begin{figure*}
    \centering
    \includegraphics[width=0.95\linewidth]{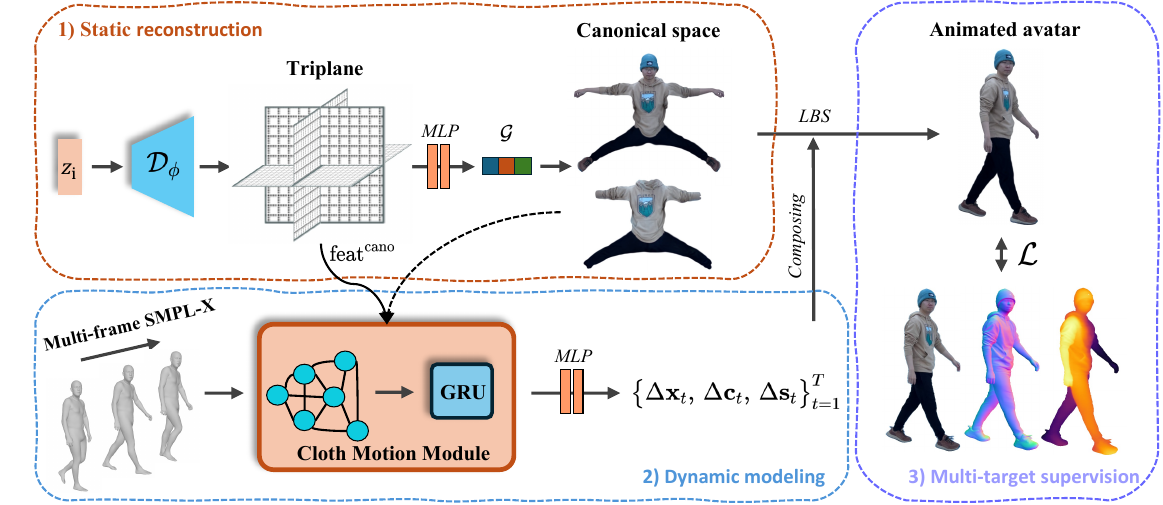}
    \caption{\textbf{MonoCloth pipeline.} \textbf{1)} We first reconstruct the static geometry and appearance in the canonical pose, where Gaussian attributes are computed and decomposed into different components. \textbf{2)} Combining static avatar features with multi-frame SMPL-X parameters, we incorporate both spatial and temporal information to predict motion-dependent offsets that enrich avatar details. \textbf{3)} The reconstructed avatar is supervised using ground-truth RGB images, normal maps, depth maps, and more auxiliary targets to jointly optimize appearance and geometry.}
    \label{fig:pipeline}
    % \vspace{-10pt}
\end{figure*}

\section{Method}

\subsection{Decomposed Clothed Human Avatar}

% Different parts of the human avatar pose distinct challenges in reconstruction and animation—for example, capturing subtle facial expressions, detailed hand movements, or the complex deformations of flexible garments. To address these challenges, we decompose the unified human avatar into functionally distinct components and apply targeted optimization strategies to each. Inspired by ExAvatar, for the face and hands, we leverage FLAME for facial geometry and joint-based optimization for elaborate hand articulation—to enable more robust and detailed reconstruction of these regions.
Different parts of the human avatar present distinct challenges for reconstruction and animation, such as modeling subtle facial expressions, fine-grained hand articulations, and the complex deformations of flexible garments. 
To effectively address these challenges, we decompose the human avatar into functionally distinct components and apply dedicated optimization strategies to each. Inspired by ~\cite{zheng2023avatarrex} and ~\cite{moon2024expressive}, we use FLAME~\cite{li2017flame} to reconstruct facial geometry and adopt joint-based optimization to capture detailed hand articulations, enabling more accurate and robust reconstruction of these critical regions.

On the other hand, clothing poses a significant challenge for modeling due to its diverse styles, soft materials, and non-rigid deformation patterns, which lead to substantial variations in geometry and appearance. 
% On the other hand, clothing presents a particularly difficult modeling problem due to its diverse styles, soft materials, and non-rigid deformation patterns, which lead to significant variations in geometry and appearance.
% These effects are difficult to capture simply with parametric models and rigid articulations. 
These effects are difficult to capture using parametric models and rigid articulations.
To enable more realistic reconstruction, we decouple clothing from the body and model it independently.
Given the limited 3D cues in monocular videos, we adopt SMPL-X~\cite{pavlakos2019smplx} as a coarse 3D prior. We place one 3D Gaussian at each upsampled SMPL-X mesh vertex and assign each to a semantic component. Formally, the complete set of Gaussians $\mathcal{G}$ is decomposed as:

\begin{equation}
\mathcal{G} = \mathcal{G}^{\text{face}} \cup \mathcal{G}^{\text{hands}} \cup \mathcal{G}^{\text{cloth}} \cup \mathcal{G}^{\text{body}},
\end{equation}

\noindent where $\mathcal{G}^{\text{face}}$, $\mathcal{G}^{\text{hands}}$, and $\mathcal{G}^{\text{cloth}}$ denote the Gaussians associated with the face, hands, and clothing, respectively. The remaining Gaussians are assigned to $\mathcal{G}^{\text{body}}$.

Specifically, $\mathcal{G}^{\text{face}}$ and $\mathcal{G}^{\text{hands}}$ are obtained through vertex correspondences between SMPL-X and the FLAME/MANO~\cite{romero2022mano} models, while $\mathcal{G}^{\text{cloth}}$ is derived by 2D clothing segmentations. These 2D segmentations serve as pseudo-ground-truth labels to train a lightweight classifier that assigns semantic labels to Gaussian points. 
To address errors caused by occlusions and limited viewpoints in monocular videos, we introduce a connectivity-based refinement step that corrects misclassifications and enforces spatial consistency. Please see the supplementary material for more details.

% This decomposition enables independent modeling and supervision of each component, leading to more accurate reconstruction and controllable animation.

\subsection{MonoCloth Pipeline}

% To bridge the gap between the limited information available in monocular videos and the complex dynamics of physical human motion, we employ a two-stage training strategy.

\textbf{Two-stage strategy.} To address the challenge of learning complex non-rigid motion dynamics from short monocular video clips, we adopt a two-stage strategy that leverages a broader collection of monocular videos to capture shared patterns of human appearance and cloth motion. To enable identity-specific representation, each subject is assigned a compact latent code $\mathbf{z}_{i} \in \mathbb{R}^{64}$.

In Stage 1, we train on monocular videos of multiple subjects to jointly reconstruct avatars of each individual. The only identity-specific component is the latent code, enabling the shared network to learn generalized priors for human geometry, appearance, and motion. Leveraging these learned priors, we then fine-tune the pretrained model on a single subject in Stage 2 to further enhance reconstruction fidelity.

\noindent \textbf{Static reconstruction.} In both training stages, avatar reconstruction begins by decoding the latent code into a triplane feature representation. Specifically, each identity $i$ is associated with a latent code, which is decoded by a shared decoder $\mathcal{D}_\phi$ parameterized by $\phi$, as follows:

\begin{equation}
\mathcal{T}_{i} = \mathcal{D}_\phi(\mathbf{z}_{i}) = \left\{ \mathcal{T}^x_{i},\ \mathcal{T}_i^{y},\ \mathcal{T}_i^{z} \right\},
\end{equation}

\noindent where each $\mathcal{T}^*_{i} \in \mathbb{R}^{C\times H\times W}$ is a 2D feature map ($C=32$, $H=128$, $W=128$), and the set $\mathcal{T}_i$ forms a triplane encoding both geometry and appearance information.

For feature extraction, inspired by ~\cite{kocabas2024hugs} and ~\cite{moon2024expressive}, we use an upsampled SMPL-X mesh as the geometric base. To eliminate shape variation across subjects, we set all identity-dependent SMPL-X shape parameters to zero, resulting in a standard reference mesh in the canonical space, denoted as $\mathbf{v}^{\text{cano}}$.
Given the vertices of $\mathbf{v}^{\text{cano}}$, we extract per-vertex features $\text{feat}^\text{cano}$ by bilinearly sampling across the three orthogonal 2D planes. For region-specific features such as $\text{feat}^\text{cloth}$, we select the corresponding vertices based on the Gaussian set $\mathcal{G}^{*}$ of each component.
% Given the vertices of $\mathbf{v}_{\text{std}}$, we extract per-vertex features $\mathbf{f}_v$ by bilinearly sampling across the three orthogonal 2D planes. For region-specific features like $\mathbf{f}_{\text{cloth}}$, we select the corresponding vertices based on the 3D segmentation $\mathcal{G}_{*}$ of each component.

% The extracted features are fed into geometry and appearance decoder MLPs to predict per-vertex Gaussian attributes, including the spatial displacement $\Delta \textbf{x}$ from the subject-specific SMPL-X mesh $\mathbf{v}_{\text{id}}$, the base color $\textbf{c}$, and the scale $\textbf{s}$. 
% The resulting mesh in the canonical pose is thus given by $\mathbf{v}_{\text{cano}}=\mathbf{v}_{\text{id}}+\Delta \textbf{x}$.
% By constraining all Gaussians to be isotropic and setting their opacity to 1, we obtain the final Gaussian set $\mathcal{G}=\{\textbf{x},\textbf{c},\textbf{s}\}$,
% which can be rendered using Gaussian Splatting to reconstruct the avatar’s static geometry and appearance in canonical space.

The extracted features are passed through geometry and appearance decoder MLPs to predict per-vertex Gaussian attributes, including the spatial displacement $\Delta \mathbf{x}$ from the subject-specific SMPL-X mesh $\mathbf{v}^{\text{id}}$, base color $\mathbf{c}$, and scale $\mathbf{s}$.
The identity-dependent mesh in the canonical pose is then computed as $\mathbf{v}^{\text{id}} = \mathbf{v}^{\text{cano}} + \Delta \mathbf{x}$.
% Assuming isotropic Gaussians with fixed opacity of 1, we construct the final Gaussian set $\mathcal{G} = \{\mathbf{x}, \mathbf{c}, \mathbf{s}\}$, which is rendered via Gaussian Splatting to reconstruct the avatar’s static geometry and appearance in canonical space.
Following prior works~\cite{hu2024gaussianavatar}, we assume isotropic Gaussians with fixed opacity of 1 to mitigate the limited information inherent in monocular videos. The final Gaussian set $\mathcal{G} = \{\mathbf{x}, \mathbf{c}, \mathbf{s}\}$ is rendered via Gaussian Splatting to reconstruct the avatar’s static geometry and appearance in canonical space.

% \noindent \textbf{Dynamic modeling.} We then simulate the motion-dependent variations of the human avatar. For exposed body regions such as arms and lower legs, which exhibit relatively minor deformations during movement, we adopt a pose-dependent module to estimate small displacement deltas. In contrast, clothing regions undergo complex, non-rigid deformations due to their flexible material properties and motion history. To model this more accurately, we introduce a dedicated cloth motion simulation module that incorporates both spatial and temporal information.

% \noindent \textbf{Dynamic modeling.} We simulate motion-dependent variations of the human avatar by separately modeling the dynamics of different body regions. For exposed body parts such as arms and lower legs, which exhibit relatively minor non-rigid deformation, we employ a pose-dependent module to estimate small displacement offsets. In contrast, clothing regions experience complex, non-rigid deformations due to their flexible material properties and the influence of motion history. To more accurately capture these effects, we introduce a cloth motion simulation module (\textbf{CloSim}) that incorporates both spatial and temporal information.

\noindent \textbf{Dynamic modeling.} To simulate motion-dependent variations, we model the dynamics of different body regions independently. For exposed body parts such as arms and lower legs, which undergo relatively minor non-rigid deformation, we use a pose-dependent module to predict small displacement offsets. In contrast, clothing exhibits complex, non-rigid deformations driven by flexible material properties and motion history. To better capture these dynamics, we introduce the cloth motion simulation module (\textbf{CloSim}) that integrates both spatial and temporal information.

% For the input to CloSim, we utilize the 3D clothing segmentation of the avatar to extract features from the corresponding region.  
% These features are then combined with a sequence of SMPL-X poses from multiple frames, enabling the input to capture both geometric context and temporal dynamics.
For the input to Closim, we combine the clothing feature $\text{feat}^{\text{cano}}$ from the canonical space with a sequence of SMPL-X poses from multiple frames, enabling the input to capture both geometric context and temporal dynamics.
To ensure sufficient motion variation, frames are sampled at a low frequency (i.e., 5 FPS), with one frame before and one after the current time step included to provide bidirectional temporal context. MonoCloth adopts a temporal window as:

\begin{equation}
\mathcal{P}_T = \left\{ \boldsymbol{\theta}_{T - \Delta t},\ \boldsymbol{\theta}_{T},\ \boldsymbol{\theta}_{T + \Delta t} \right\},
\end{equation}

% \noindent where $\boldsymbol{\theta}$ denotes the SMPL-X pose parameters and $\Delta t$ is set to 0.2.
\noindent where $\boldsymbol{\theta}$ denotes the SMPL-X pose parameters, and $\Delta t$ represents the temporal offset, set to 0.2 seconds.

Through CloSim, we obtain per-time-step offsets that encode motion dynamics relative to the canonical avatar.

\noindent \textbf{Multi-target supervision.}
% After obtaining the motion offset, we adopt a deformation strategy that first applies the offset and then performs Linear Blend Skinning (LBS). This approach has the advantage that the offset is defined in the canonical pose space, enabling the model to share generalizable motion knowledge across different individuals without being affected by the variations introduced by LBS. The deformation process is formulated as:
After obtaining the motion offsets, we adopt a deformation strategy that first applies the offsets to the canonical pose and then performs Linear Blend Skinning (LBS). This design ensures that the predicted offsets are defined independently of articulation, allowing the model to learn generalizable motion patterns across individuals without being affected by the transformations introduced by LBS. The deformation process is defined as:

\begin{equation}
\mathbf{v}_t = \text{LBS}(\mathbf{v}^{\text{id}} + \Delta \mathbf{x}_t,\ \boldsymbol{\theta}_t),
\end{equation}

\noindent where $\mathbf{v}^{\text{id}}$ is the mesh in the canonical pose and $\Delta \mathbf{x}_t$ denotes the predicted per-vertex offset at time $t$.

% To enhance the model's 3D perception, we compute per-vertex normals $\mathbf{n}_t$ and depth values $\mathbf{d}_t$ from the deformed mesh $\mathbf{v}_t$. These are rendered into 2D normal and depth maps, denoted as $\hat{\mathbf{N}}_t$ and $\hat{\mathbf{D}}_t$, via Gaussian Splatting. The rendered maps are supervised by corresponding ground-truth maps $\mathbf{N}_t^{\text{gt}}$ and $\mathbf{D}_t^{\text{gt}}$, obtained from the vision foundation model SAPIENS.
% This additional normal and depth supervision enhance the geometric accuracy of the reconstructed avatar and improve the fidelity of motion-dependent surface detail.

To enhance the model’s 3D perception, we compute per-vertex normals $\mathbf{n}_t$ and depth values $\mathbf{d}_t$ from the deformed mesh $\mathbf{v}_t$. These quantities are then rendered into 2D normal and depth maps, denoted as $\hat{\mathbf{N}}_t$ and $\hat{\mathbf{D}}_t$, using Gaussian Splatting. The rendered maps are supervised by pseudo ground-truth $\mathbf{N}_t^{\text{gt}}$ and $\mathbf{D}_t^{\text{gt}}$, generated by the vision foundation model Sapiens~\cite{khirodkar2024sapiens}.
This additional supervision improves the geometric accuracy of the reconstructed avatar and enhances the fidelity of motion-dependent surface details.

% \noindent \textbf{Multi-target supervision.} After obtaining the motion offsets, we adopt a deformation strategy similar to ExAvatar by applying the offsets before performing linear blend skinning (LBS). This approach ensures that the predicted offsets are defined in the canonical pose space, allowing generalizable motion knowledge to be shared across different identities without being entangled with LBS transformations.

% \begin{equation}
% \mathbf{v}_T = \text{LBS}(\mathbf{v}_{\text{cano}} + \Delta \mathbf{x}_T,\ \boldsymbol{\theta}_T)
% \end{equation}

% Finally, we render the avatar using Gaussian splatting. Additionally, we leverage the connectivity of the SMPL-X mesh to compute per-vertex normals and depth values, which are also rendered through the Gaussian renderer to produce normal and depth maps. These maps serve as additional supervision signals during training.

\subsection{Cloth Simulation Module}

% Clothing is one of the most challenging components to model in human motion, due to its flexible and loose nature. These physical properties lead to highly dynamic and intricate variations in both geometry and appearance across video frames. To model this effectively, we propose the Cloth Motion Simulation Module (\textbf{CloSim}) to capture the detailed motion of clothing.

% Through avatar decomposition, we can decouple the clothing regions from the full-body avatar for targeted processing. Given the rich geometric and appearance details of clothing, we introduce a dedicated feature extraction branch specifically for garments. This enables the extraction of fine-grained features to support high-fidelity reconstruction in subsequent stages.

Clothing is particularly challenging to model in human motion due to its flexible and loose nature, resulting in complex, non-rigid variations in geometry and appearance across frames. To address this, we propose CloSim to capture detailed garment dynamics.
% Leveraging avatar decomposition, we can decouple clothing regions from the full-body avatar. Then, to handle the rich geometric and appearance details of garments, we utilize a dedicated feature extraction branch that enables fine-grained representation for further reconstruction and refinement.
Leveraging avatar decomposition, we first decouple the clothing regions from the full-body avatar. Then, to capture the rich geometric and appearance details of garments, we employ a dedicated feature extraction branch, which enables fine-grained representations for subsequent reconstruction and refinement.

% \noindent\textbf{Spatio-temporal modeling.} To capture the dynamics of clothing during avatar motion, we model the overall deformation as the collective effect of individual movements of clothing-related Gaussian points. Given the spatial continuity of clothing geometry, the motion of each Gaussian point is naturally influenced by its neighbors. Therefore, it is critical to model the interactions among these points. To achieve this, we employ a Graph Convolutional Network (GCN), which facilitates information propagation among neighboring Gaussian points. This allows each point to incorporate local deformation context and better capture collective behavior in cloth simulation:

\noindent\textbf{Spatio-temporal modeling.} To capture cloth dynamics during avatar motion, we model overall deformation as the collective behavior of clothing-related Gaussian points $\mathcal{G}^{\text{cloth}}$. Due to the spatial continuity of cloth geometry, the motion of each point is influenced by its neighbors, making it essential to model their interactions. To this end, we employ a Graph Convolutional Network (GCN)~\cite{kipf2016semi} that enables information propagation across connected points, allowing each point to incorporate local deformation context and better capture collective motion patterns:

\begin{equation}
    Z_t = \mathrm{GCN}\bigl(\text{Concat}(\,\text{feat}^{\text{cano}}, \text{feat}^{\text{pose}}_t),\,\mathcal{E}\bigr),
\end{equation}

\noindent where $Z^t\in\mathbb{R}^{N\times d}$ denotes the encoded features for all Gaussians, each with dimension $d= 128$. The GCN processes the concatenation of the canonical-space features $\text{feat}^{\text{cano}}\in\mathbb{R}^{N\times96}$ and body pose feature $\text{feat}^{\text{pose}}_t\in\mathbb{R}^{N\times126}$, using the mesh connectivity $\mathcal{E}$ to guide message passing.

% Monocular avatar reconstruction methods typically rely only on body pose to drive deformation. However, this is insufficient for modeling the complex behavior of garments. Cloth dynamics are affected not only by body pose but also by inertia, gravity, and other physical effects, making temporal continuity essential. To capture such sequential dependencies, we integrate a Gated Recurrent Unit (GRU), which allows the network to track motion trajectories and evolving garment states over time:

Monocular avatar reconstruction methods typically rely solely on body pose to drive deformation. However, this is insufficient for modeling the complex dynamics of garments, which are influenced not only by pose but also by inertia, gravity, and other physical factors. Capturing temporal continuity is therefore essential. To model these sequential dependencies, we integrate a Gated Recurrent Unit (GRU)~\cite{chung2014empirical}, enabling the network to track motion trajectories and evolving garment states over time:

\begin{equation}
\bigl\{ \Delta \mathbf{x}_t,\, \Delta \mathbf{c}_t,\, \Delta \mathbf{s}_t \bigr\}_{t=1}^T
= \Psi\!\left(\mathrm{GRU}(\{Z_t\}_{t=1}^T,\ h_0)\right),
\end{equation}

\noindent where $\Delta \mathbf{x}_t \in \mathbb{R}^3$, $\Delta \mathbf{c}_t\in \mathbb{R}^3$, and $\Delta \mathbf{s}_t\in \mathbb{R}$ represent the predicted residuals for position, color, and scale of each Gaussian point at time $t$, respectively. 
% The GRU takes as input the sequence of encoded features $\{Z_t\}_{t=1}^T$ along with an initial hidden state $h^0$, and outputs temporally-aware embeddings, which are decoded by a lightweight MLP $\Psi$. 
Given the sequence of encoded features $\{Z_t\}_{t=1}^T$ and an initial hidden state $h_0$, the GRU generates temporally-aware embeddings, which are subsequently decoded by a lightweight MLP $\Psi$.
The output lies in $\mathbb{R}^{T \times N \times 7}$, representing per-frame offsets for all $N$ points across $T$ frames.

% \begin{figure}
%     \centering
%     \includegraphics[width=0.9\linewidth]{AnonymousSubmission/LaTeX/figures/cloth_module.pdf}
%     \caption{Cloth motion module.}
%     \label{fig:enter-label}
% \end{figure}

\noindent \textbf{Sampling strategy.} To improve the model’s robustness to temporal variation, we adopt a data augmentation strategy based on random supervision during training. Specifically, while CloSim consistently receives inputs at $\mathcal{P}_T$  and outputs motion offsets $\{\Delta \mathbf{x}_t,\,\Delta \mathbf{c}_t,\,\Delta\mathbf{s} _t\}$ for the frames $\{T-\Delta t,\ T,\ T+\Delta t\}$, supervision is applied only at randomly sampled time points $t \in [T-\Delta t,\ T+\Delta t]$, with the corresponding offset computed via linear interpolation as:

\begin{equation}
\Delta \hat{\mathbf{x}}_t = (1 - \alpha) \Delta \mathbf{x}_{T - \Delta t} + \alpha \Delta \mathbf{x}_{T + \Delta t},
\end{equation}

\noindent where $\alpha$ represents the normalized position of $t$ within the interval. The offsets $\Delta \hat{\mathbf{c}}_t$ and $\Delta \hat{\mathbf{s}}_t$ are computed similarly.

\input{AnonymousSubmission/LaTeX/fig_tex/neuman}

\subsection{Loss Functions}

% \textbf{Rendering loss.} We employ a combination of L1 loss $\mathcal{L}_{\text{rgb}}$, structural similarity loss $\mathcal{L}_{\text{ssim}}$, and and perceptual loss $\mathcal{L}_{\text{lpips}}$ to supervise the rendered images $I_{\text{pred}}$ against the ground truth $I_{\text{gt}}$. This encourages both pixel-wise accuracy and perceptual fidelity. To further enhance the appearance reconstruction of clothing regions, we introduce an additional loss term $\mathcal{L}_{\text{cloth}}$ between the rendered 3D clothing and the masked 2D clothing image to better capture the fine-grained appearance details of garments.

\textbf{Rendering loss.} To supervise the rendered images $I_{\text{pred}}$ against the ground truth $I_{\text{gt}}$, we employ a combination of L1 loss $\mathcal{L}_{\text{rgb}}$, structural similarity loss $\mathcal{L}_{\text{ssim}}$, and perceptual loss $\mathcal{L}_{\text{lpips}}$, encouraging both pixel-level accuracy and perceptual fidelity.
To further enhance appearance reconstruction in clothing regions, we introduce an additional loss term $\mathcal{L}_{\text{cloth}}$, computed between the rendered 3D clothing and the masked 2D clothing image, to better capture fine-grained garment details.

% \noindent\textbf{Geometry loss.} To better capture the fine-grained dynamics of cloth motion, we introduce dedicated supervision on the 3D geometry. Based on the deformed 3D avatar mesh, we compute the vertex normals and convert them into RGB values, treating these as the color of the corresponding Gaussian points. We then render a normal map of the generated avatar and compute a cosine similarity loss against the normal map predicted by SAPIENs. Similarly, we render the depth map of the avatar and and use L1 loss to supervise it using the depth map from SAPIENs. Lastly, to enhance geometric alignment along clothing boundaries, we incorporate a silhouette loss to enforce edge consistency and improve the accuracy of cloth contours. The final geometry loss is formulated as:

% \begin{equation}
% \mathcal{L}_{\text{geo}} = \lambda_{\text{normal}} \mathcal{L}_{\text{normal}} + \lambda_{\text{depth}} \mathcal{L}_{\text{depth}} + \lambda_{\text{silhouette}} \mathcal{L}_{\text{silhouette}},
% \end{equation}

% \noindent where $\lambda_{\text{normal}}=5$, $\lambda_{\text{depth}}=1$, and $\lambda_{\text{silhouette}}=2$.

\noindent\textbf{Geometry loss.}
To capture the fine-grained dynamics of cloth motion, we introduce explicit supervision on 3D geometry. Given the deformed mesh, we compute vertex normals and encode them as RGB values for the corresponding Gaussian points. 
The rendered normal map $\mathbf{N}_{\text{pred}}$ is compared with the reference map $\mathbf{N}_{\text{gt}}$ from Sapiens~\cite{khirodkar2024sapiens} using a cosine similarity loss: $\mathcal{L}_{\mathbf{N}} = 1 - \langle \mathbf{N}_{\text{pred}}, \mathbf{N}_{\text{gt}} \rangle$. Similarly, the depth map $\mathbf{D}_{\text{pred}}$ is supervised using an L1 loss against the ground truth $\mathbf{D}_{\text{gt}}$: $\mathcal{L}_{\mathbf{D}} = \| \mathbf{D}_{\text{pred}} - \mathbf{D}_{\text{gt}} \|_1$. To improve boundary alignment, we also apply a silhouette loss between predicted and ground-truth masks: $\mathcal{L}_{\mathbf{S}} = \| \mathbf{S}_{\text{pred}} - \mathbf{S}_{\text{gt}} \|_2^2$.
The final geometry loss is defined as:

\begin{equation}
    \mathcal{L}_{\text{geo}} = \lambda_{\mathbf{N}} \mathcal{L}_{\mathbf{N}} + \lambda_{\mathbf{D}} \mathcal{L}_{\mathbf{D}} + \lambda_{\mathbf{S}} \mathcal{L}_{\mathbf{S}},
\end{equation}

\noindent where the weights are set to $\lambda_{\mathbf{N}} = 5$, $\lambda_{\mathbf{D}} = 1$, and $\lambda_{\mathbf{S}} = 2$.

\noindent\textbf{Temporal loss.}
To enforce temporal coherence across frames, we introduce a temporal consistency loss that penalizes temporal discontinuities in geometry, color, and scale offsets between consecutive frames. Specifically, let $\Delta \mathbf{x}_t$, $\Delta \mathbf{c}_t$, and $\Delta \mathbf{s}_t$ denote the predicted offsets at frame $t$. The temporal loss is defined as:

\begin{equation}
\begin{aligned}
\mathcal{L}_{\text{temp}} 
&= \lambda_{\text{temp}} \sum_{t=1}^{T-1} \big( 
\| \Delta \mathbf{x}_{t+1} - \Delta \mathbf{x}_t \|_2^2 \\
&\quad + \| \Delta \mathbf{c}_{t+1} - \Delta \mathbf{c}_t \|_2^2 
+ \| \Delta \mathbf{s}_{t+1} - \Delta \mathbf{s}_t \|_2^2 \big),
\end{aligned}
\end{equation}

\noindent where $\lambda_{\text{temp}} = 0.1$ controls the strength of temporal regularization. This encourages smooth transitions over time and reduces artifacts in cloth animation.

\noindent\textbf{Final loss.}
% In addition to the core losses described above, we incorporate several auxiliary loss terms to enhance training stability and reconstruction accuracy. Inspired by ExAvatar and GaussianAvatar, we introduce regularization losses such as L2-norm penalties to stabilize the optimization of geometry. Furthermore, for facial and hand regions, we leverage vertex-level supervision based on parametric models to constrain the 3D positions of the corresponding Gaussian points, enhancing the robustness of the geometry reconstruction of the face and hands. Combining all the above losses, we obtain the final loss $\mathcal{L}$ used for training. For more details on the loss terms, please refer to the supplementary material.
In addition to the core loss terms described above, we incorporate several auxiliary losses to improve training stability and reconstruction accuracy. 
Inspired by ~\cite{moon2024expressive}, we introduce regularization terms such as L2-norm penalties on the predicted offsets to stabilize geometry optimization.
For facial and hand regions,  $\mathcal{G}^\text{face}$ and $\mathcal{G}^\text{hands}$, we further apply vertex-level supervision based on parametric models, directly constraining the 3D positions of the associated Gaussian points. 
This enhances the robustness and precision of geometry reconstruction in these high-detail areas.
All loss components are combined to form the final training objective $\mathcal{L}$. Please refer to the supplementary material for more details.

%% file: AnonymousSubmission/LaTeX/fig_tex/neuman.tex
\begin{figure*}[t]
    \centering
    \includegraphics[width=0.75\linewidth]{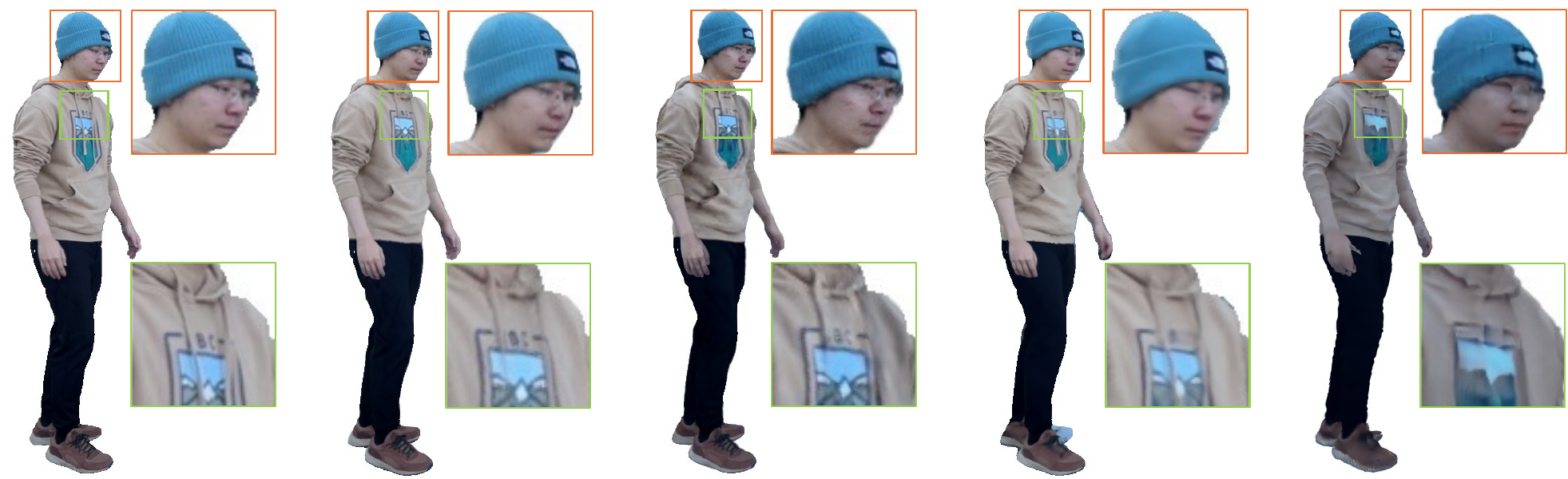}\hspace{5pt}
    \\
    \includegraphics[width=0.75\linewidth]{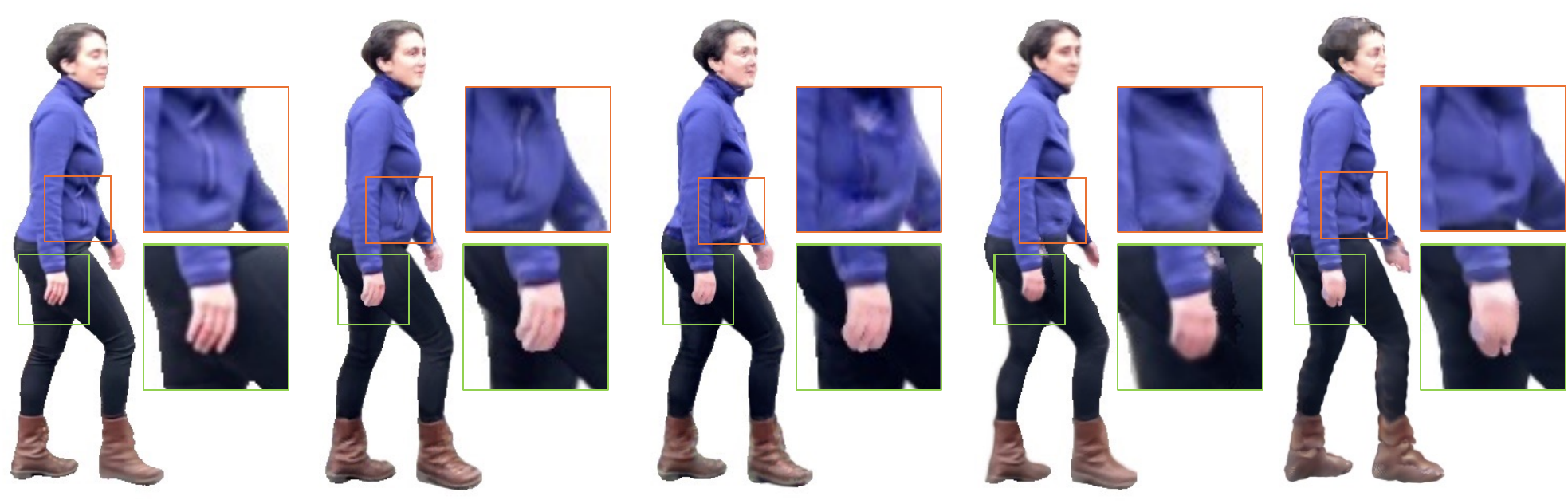}\hspace{5pt}
    \\
    \makebox[0.000\linewidth]{}\hspace{5pt}
    \makebox[0.13\linewidth]{\footnotesize\textbf{GT}}\hspace{5pt}
    \makebox[0.00\linewidth]{}\hspace{0pt}
    \makebox[0.13\linewidth]{\footnotesize\textbf{Ours}}\hspace{5pt}
    \makebox[0.00\linewidth]{}\hspace{5pt}
    \makebox[0.13\linewidth]{\footnotesize\textbf{ExAvatar}}\hspace{5pt}
    \makebox[0.00\linewidth]{}\hspace{5pt}
    \makebox[0.13\linewidth]{\footnotesize\textbf{Vid2Avatar}}\hspace{15pt}
    \makebox[0.13\linewidth]{\footnotesize\textbf{GaussianAvatar}}\hspace{5pt}
    \makebox[0.03\linewidth]{}\hspace{5pt}
    \caption{\textbf{Qualitative results on NeuMan.} Our method achieves the highest overall visual quality, particularly in reconstructing fine clothing textures as well as facial and hand details.}
    \label{fig:neuman}
\end{figure*}

%% file: AnonymousSubmission/LaTeX/secs/5_experiments.tex
\section{Experiments}

\subsection{Experimental Setups}

% We conduct training and evaluation using two monocular human video datasets: NeuMan and XHuman. The NeuMan dataset contains outdoor motion videos of six individuals, with frames subsampled to increase pose variation between adjacent frames. Due to its low frame rate and the use of non-consecutive frames in the test set, we treat NeuMan as a benchmark for novel pose generalization.
\textbf{Datasets and metrics.}
We conduct training and evaluation on two monocular human video datasets: NeuMan~\cite{jiang2022neuman} and X-Humans~\cite{shen2023xhuman}. The NeuMan dataset contains in-the-wild motion videos of six individuals, with frames down-sampled to increase pose variation between adjacent frames. 
Given its low frame rate and non-consecutive testing frames, NeuMan serves as a benchmark for evaluating avatar reconstruction and generalization to novel poses.
In contrast, X-Humans provides a larger collection of high-quality monocular videos captured in controlled laboratory settings. It contains motion data from 20 subjects, along with corresponding RGB-D videos, 3D meshes, and SMPL parameters. In our experiments, we only use the RGB videos of 10 subjects as input to evaluate the model’s performance on temporally continuous video sequences.

% In contrast, XHumans offers a larger collection of high-quality monocular videos captured in lab environments. It includes data from 20 subjects in motion, along with corresponding RGB-D videos, 3D meshes, and SMPL parameters. For our experiments, we mainly use the RGB videos as input to evaluate our model’s performance on temporally continuous video sequences.

We evaluate the rendering quality of all methods using PSNR, SSIM, and LPIPS~\cite{zhang2018unreasonable} metrics. Since the NeuMan dataset contains outdoor backgrounds, we apply a mask to replace the background with a white canvas before the evaluation. This ensures that the evaluation focuses solely on the quality of the avatar reconstruction.

\noindent\textbf{Implementation details.} We use only five subjects from the X-Humans dataset with large appearance and motion variations for Stage 1 training, aiming to assess the potential of pretraining to enhance model performance. The model is optimized using the Adam optimizer with an initial learning rate of $10^{-3}$. On a single NVIDIA RTX 4090 GPU, our method takes approximately 5 minutes to complete 3D clothing segmentation and around 3.5 hours to reconstruct a single subject from the NeuMan dataset on average. We provide more details in the supplementary material.

\begin{table}[ht]
  \centering
  
  \begin{tabular}{lccc}
    \toprule
    Method                  & PSNR $\uparrow$       & SSIM $\uparrow$        & LPIPS $\downarrow$     \\
    \midrule
    HumanNeRF     & 27.06                 & 0.967                  & 1.92                   \\
    InstantAvatar & 28.47                 & 0.972                  & 2.77                   \\
    % NeuMan        & 25.48                 & 0.966                  & 2.87                   \\
    Vid2Avatar    & 29.48                 & 0.976                  & 1.85                   \\
    3DGS-Avatar   & 29.75                 & 0.975                  & 1.75                   \\
    GaussianAvatar& 28.90 & 0.974 & 1.81 \\
    ExAvatar      & 31.70 & 0.982 & 1.47 \\
    % $^*$PGHM      & 31.85 & 0.987 & 1.71 \\
    $^*$Vid2Avatar-Pro       &    32.71 &    0.983 &    \textbf{1.19} \\
    \midrule
    Ours (w/o pretrain)       &    \underline{33.18} &    \underline{0.985} &    1.28 \\
    Ours (full)  &    \textbf{33.53} &    \textbf{0.986} &    \underline{1.20} \\
    \bottomrule
  \end{tabular}
  \caption{\textbf{Quantitative results on NeuMan.} LPIPS is measured on the scale of $10^{-2}$. The best and second-best results are highlighted in boldface and underlined, respectively.}
  \label{tab:neuman}
\end{table}

\subsection{Evaluation and Comparison}

We first evaluate the quality of avatar reconstruction on the NeuMan dataset. Following previous works~\cite{hu2024gaussianavatar}, we use four high-quality sequences (e.g., Seattle, Bike, Citron, and Jogging) for quantitative evaluation. Tab.~\ref{tab:neuman} presents comparisons between our method and several state-of-the-art approaches. 
Some results are partially sourced from ~\cite{hu2024gaussianavatar} and ~\cite{guo2025vid2avatarpro}. 
While all methods reconstruct avatars from monocular videos, Vid2Avatar-Pro~\cite{guo2025vid2avatarpro} leverages pretraining on a large-scale multi-view dataset, providing stronger 3D priors at the cost of increased data collection complexity. To highlight this distinction, we mark it with a star in Tab.~\ref{tab:neuman}.
To ensure a fair comparison, we report both the results of our full method and an ablated version without any pretrained model. Benefiting from our part-based optimization design, our approach achieves state-of-the-art performance even without monocular video pretraining. Fig.~\ref{fig:neuman} visualizes the reconstruction and animation results, showing that our method produces the highest overall quality, especially in capturing detailed clothing textures as well as facial and hand details.

\begin{table}[ht]
  \centering
  \begin{tabular}{lccc}
    \toprule
    Method                  & PSNR $\uparrow$       & SSIM $\uparrow$        & LPIPS $\downarrow$     \\
    \midrule
    ExAvatar      & \underline{29.41} & \underline{0.973} & \underline{2.24} \\

    Ours  &    \textbf{30.68} &    \textbf{0.976} &    \textbf{2.21} \\
    \bottomrule
  \end{tabular}
  \caption{\textbf{Quantitative results on X-Humans.} Our method achieves superior performance in video animation quality.}
  \label{tab:xhuman}
\end{table}

% \begin{figure}[htbp]
%     % \vspace{-10pt}
%     \centering
%     \includegraphics[width=0.65\linewidth]{AnonymousSubmission/LaTeX/figures/xhuman.png}\hspace{5pt}
%     \\
%     \makebox[0.02\linewidth]{}\hspace{5pt}
%     \makebox[0.2\linewidth]{\footnotesize\textbf{GT}}\hspace{5pt}
%     \makebox[0.2\linewidth]{\footnotesize\textbf{ExAvatar}}\hspace{5pt}
%     \makebox[0.2\linewidth]{\footnotesize\textbf{Ours}}\hspace{5pt}
%     \makebox[0.02\linewidth]{}\hspace{5pt}
%     \caption{\textbf{Comparison on XHuman.} Benefiting from our temporal modeling, clothing textures remain more stable during large movements.}
%     % \vspace{-10pt}
%     \label{fig:xhuman}
% \end{figure}

\begin{figure}[htbp]
  \centering
  % \vspace{-5pt}
  \begin{minipage}[t]{0.48\linewidth}
    \centering
    \includegraphics[width=\linewidth]{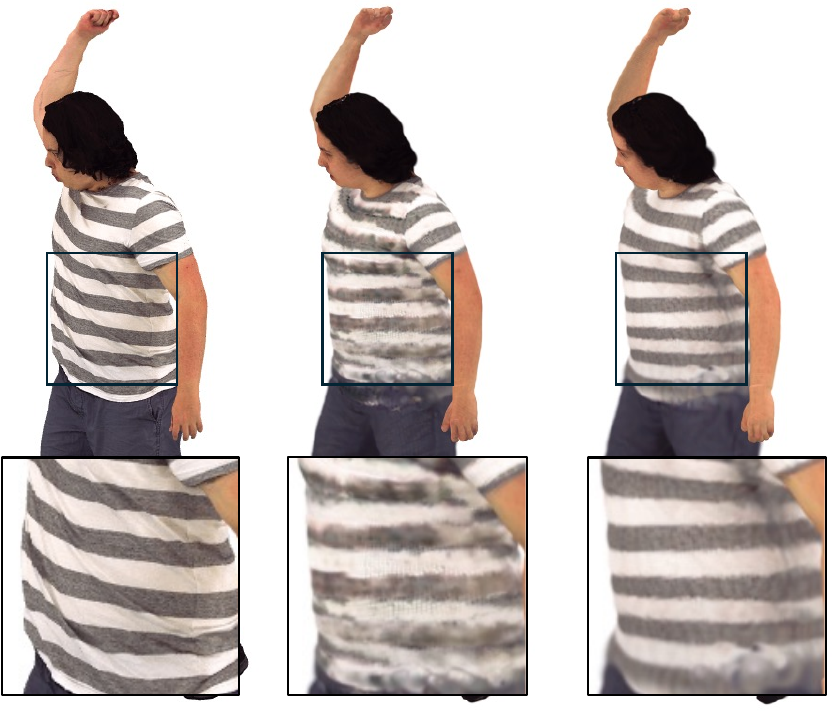}
    
    \vspace{2pt}
    % \hspace{-2pt}
    \makebox[0.28\linewidth]{\footnotesize\textbf{GT}}
    % \hspace{2pt}
    \makebox[0.02\linewidth]{}
    \makebox[0.28\linewidth]{\footnotesize\textbf{ExAvatar}}
    \makebox[0.02\linewidth]{}
    \makebox[0.28\linewidth]{\footnotesize\textbf{Ours}}
    \makebox[0.01\linewidth]{}

    \caption{\textbf{X-Humans comparison.} Temporal modeling improves clothing stability.}
    \label{fig:xhuman}
  \end{minipage}
  \hfill
  \begin{minipage}[t]{0.45\linewidth}
    \centering
    \includegraphics[width=\linewidth]{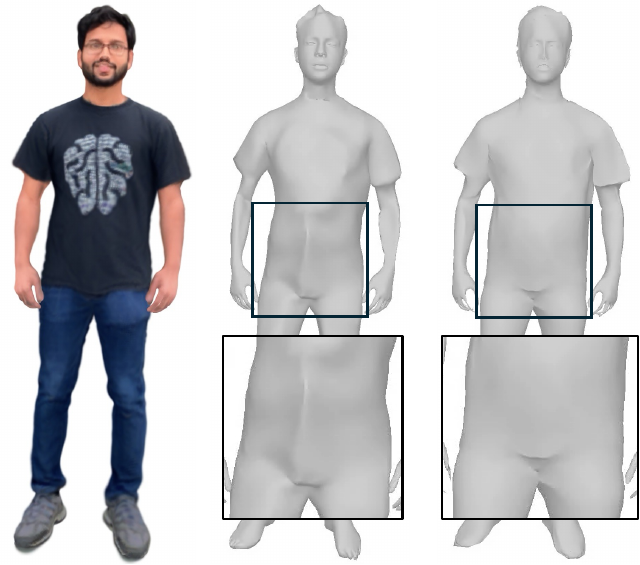}
    
    \vspace{2pt}
    \makebox[0.29\linewidth]{\footnotesize\textbf{Render}}
    \makebox[0.02\linewidth]{}
    \makebox[0.29\linewidth]{\footnotesize\textbf{w/o $\mathcal{L}_{\text{geo}}$}}
    \makebox[0.29\linewidth]{\footnotesize\textbf{w/ $\mathcal{L}_{\text{geo}}$}}
    \makebox[0.01\linewidth]{}

    \caption{\textbf{Geometry loss ablation.} Geometry supervision reduces 3D artifacts.}
    \label{fig:geometry_loss}
  \end{minipage}
  % \vspace{-5pt}
\end{figure}

Tab.~\ref{tab:xhuman} presents the evaluation results on the X-Humans dataset, where we compare our method against ExAvatar~\cite{moon2024expressive}, which is the strongest baseline on the NeuMan dataset. Our method consistently outperforms ExAvatar across all metrics. Since both the training and testing sequences in X-Humans are temporally continuous, modeling clothing dynamics over time becomes especially critical. As shown in Fig.~\ref{fig:xhuman}, ExAvatar suffers from noticeable artifacts when handling drastic motions, while our method demonstrates superior robustness and temporal stability.

% \begin{figure}[htbp]
%     % \vspace{-10pt}
%     \centering
%     \includegraphics[width=0.65\linewidth]{AnonymousSubmission/LaTeX/figures/normal.pdf}\hspace{5pt}
%     \\
%     \makebox[0.02\linewidth]{}\hspace{5pt}
%     \makebox[0.2\linewidth]{\footnotesize\textbf{Render}}\hspace{5pt}
%     \makebox[0.2\linewidth]{\footnotesize\textbf{w/o $\mathcal{L}_{\text{geo}}$}}\hspace{5pt}
%     \makebox[0.2\linewidth]{\footnotesize\textbf{w/ $\mathcal{L}_{\text{geo}}$}}\hspace{5pt}
%     \makebox[0.02\linewidth]{}\hspace{5pt}
%     \caption{\textbf{Ablation of geometry loss.} Geometry supervision mitigates 3D reconstruction artifacts.}
%     % \vspace{-10pt}
%     \label{fig:geometry_loss}
% \end{figure}

\subsection{Ablation Studies}

We conduct ablation studies on several key components of our method to assess their individual contributions. 
The results are summarized in Tab.~\ref{tab:ablation}, where we evaluate the impact of CloSim, temporal modeling, and geometry supervision.
% When motion offsets are removed and deformation is driven solely by LBS in the canonical space, many motion-dependent details are lost, leading to a significant drop in reconstruction quality. 
Removing motion offsets $\Delta_*$ from CloSim and relying solely on LBS in canonical space leads to the loss of motion-dependent details and a significant drop in reconstruction quality.
Without temporal sampling (i.e., multi-frame sampling strategy), the model cannot effectively capture temporal context, resulting in degraded rendering quality and animation smoothness.
Fig.~\ref{fig:geometry_loss} illustrates the effect of the geometry loss. 
% Without geometry supervision, due to the inherent lack of 3D information in monocular video, this can lead to incorrect estimation of clothing geometry. 
Without geometry supervision, the inherent lack of 3D information in monocular video can lead to inaccurate estimation of clothing geometry.
While the model may overfit the training views to produce visually plausible results, it often results in distorted or implausible geometry when rendered from novel viewpoints.

\begin{table}[ht]
  \centering
  \begin{tabular}{lccc}
    \toprule
    Method                  & PSNR $\uparrow$       & SSIM $\uparrow$        & LPIPS $\downarrow$     \\
    \midrule
    w/o $\Delta_*$      & 31.93 & 0.983 & 1.40 \\
    w/o temporal      & \underline{33.00} &\underline{0.985} & \underline{1.29} \\
    w/o $\mathcal{L}_{\text{geo}}$      & 32.55 & 0.985 & 1.30 \\

    Ours  &    \textbf{33.53} &    \textbf{0.986} &    \textbf{1.20} \\
    \bottomrule
  \end{tabular}
  \caption{\textbf{Ablation studies}. Evaluation on the impact of motion offsets $\Delta_*$, temporal sampling, and geometry loss $\mathcal{L}_{\text{geo}}$.}
  \label{tab:ablation}
\end{table}

\subsection{Applications}

\begin{figure}[htbp]
    % \vspace{-10pt}
    \centering
    \includegraphics[width=0.65\linewidth]{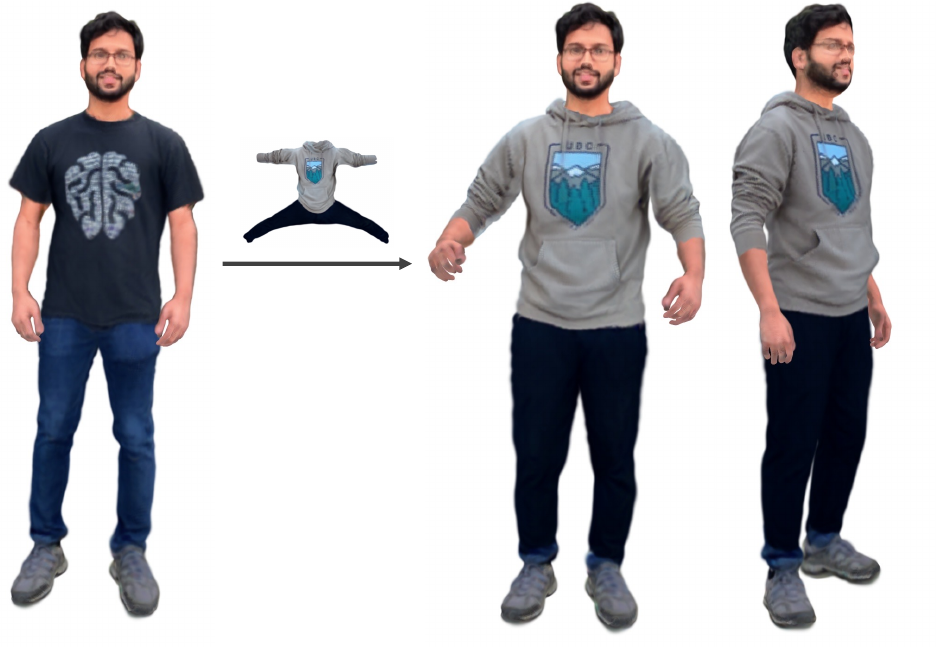}\hspace{5pt}
    \\
    \hspace{5pt}
    \makebox[0.25\linewidth]{\footnotesize\textbf{Origin}}
    % \makebox[0.3\linewidth]{\footnotesize\textbf{Novel pose}}\hspace{0pt}
    \makebox[0.15\linewidth]{}
    \makebox[0.25\linewidth]{\footnotesize\textbf{New clothed avatar}}\hspace{5pt}
    \makebox[0.1\linewidth]{}\hspace{5pt}
    \caption{\textbf{Clothing transfer.} Avatars with new garments support animation and novel-view synthesis as the original.}
    % \vspace{-10pt}
    \label{fig:application}
\end{figure}

% Our part-based decomposition and targeted optimization not only improve the reconstruction quality but also enable greater flexibility for downstream applications such as avatar editing. As shown in Fig.~\ref{fig:application}, we demonstrate a virtual try-on example by replacing the clothing of the avatar in the Citron sequence with that from the Bike sequence in the NeuMan dataset. By adjusting the SMPL-X parameters and refining the garment-body attachment, we are able to transfer and re-fit garments across different subjects.
Our part-based decomposition and targeted optimization not only enhance reconstruction quality but also offer greater flexibility for downstream applications such as avatar editing. 
As illustrated in Fig.~\ref{fig:application}, we showcase a virtual try-on example by transferring garments from the subject ``Bike'' to ``Citron'' in the NeuMan dataset. 
By adjusting the SMPL-X parameters and refining the garment-body attachment, our method enables effective clothing transfer across different subjects.
Moreover, since both the body and clothing avatars in our method are animatable, the virtual try-on results can be rendered as videos, providing a compelling visual experience for dynamic clothing transfer.

%% file: AnonymousSubmission/LaTeX/secs/6_conclusion.tex
\section{Conclusion}

We propose a novel framework for reconstructing clothed human avatars from monocular videos. To handle the varying reconstruction difficulty and motion characteristics across body parts, we decompose the avatar into components and apply targeted optimization strategies. 
% For the clothing region, which involves complex non-rigid motion, we introduce a cloth motion simulation module that incorporates spatio-temporal information. 
% By adding geometric supervision, our method enables more physically plausible garment dynamics. 
For the clothing region, which involves complex non-rigid motion, we introduce the cloth simulation module that incorporates spatio-temporal information and geometric supervision to enable more physically plausible garment dynamics.
The resulting avatars support high-quality animation, novel-view synthesis, and applications such as clothing transfer, demonstrating the flexibility and generality of our approach.

% \noindent\textbf{Limitations \& future work.} Due to the limited information in monocular videos, accurately reconstructing occluded body regions and subtle garment wrinkles during animation remains challenging. To further enhance reconstruction and animation quality, future work may incorporate more advanced vision foundation models and utilize large-scale pretraining to learn more generalized human priors.